\documentclass{article}

\usepackage{microtype}
\usepackage{graphicx}
\usepackage{subfigure}
\usepackage{booktabs} 

\usepackage{hyperref}
\usepackage{url}

\usepackage{times}  
\usepackage{helvet}  
\usepackage{courier}  
\usepackage{graphicx}  
\usepackage{multirow}

\usepackage{booktabs}
\usepackage{siunitx}
\usepackage{amsfonts}
\usepackage{appendix}
\usepackage{graphicx} 

\usepackage{algorithmic}

\usepackage{booktabs}
\usepackage{wrapfig,lipsum,booktabs}
\usepackage{bm}
\usepackage{enumitem}
\usepackage{comment}
\usepackage{changes}
\usepackage{color}

\usepackage[accepted]{icml2019}

\newcommand{\eat}[1]{}      

\usepackage{amsmath,amssymb}

\usepackage{bm} 

\usepackage{xspace}

\newcommand{\proj}{\text{proj}}
\newcommand{\sign}{\text{sign}}
\newcommand{\clip}{\text{clip}}

\def \ours {$\mathcal{N}$\textsc{Attack}\xspace}

\def \numdefense {13\xspace}
\def \numvanilla {two\xspace}

\icmltitlerunning{\ours: Learning the Distributions of Adversarial Examples for an Improved  Black-Box  Attack on Deep Neural Networks}

\begin{document}

\twocolumn[
\icmltitle{\ours: Learning the Distributions of Adversarial Examples for an Improved  Black-Box Attack on Deep Neural Networks}



\icmlsetsymbol{equal}{*}

\begin{icmlauthorlist}
\icmlauthor{Yandong Li}{equal,ucf}
\icmlauthor{Lijun Li}{equal,ucf}
\icmlauthor{Liqiang Wang}{ucf}
\icmlauthor{Tong Zhang}{HKUST}
\icmlauthor{Boqing Gong}{Google}

\end{icmlauthorlist}

\icmlaffiliation{ucf}{University of Central Florida}
\icmlaffiliation{Google}{Google}
\icmlaffiliation{HKUST}{Hong Kong University of Science and Technology}

\icmlcorrespondingauthor{Yandong Li}{lyndon.leeseu@outlook.com}

\icmlcorrespondingauthor{Boqing Gong}{BoqingGo@outlook.com}

\icmlkeywords{Adversarial attack, evolution method, black-box adversarial attack}

\vskip 0.3in
]



\printAffiliationsAndNotice{\icmlEqualContribution} 

\begin{abstract}
Powerful adversarial attack methods are vital for understanding how to construct robust deep neural networks (DNNs) and  thoroughly testing defense techniques. In this paper, we propose a black-box adversarial attack algorithm that can defeat both vanilla DNNs and those generated by various defense techniques developed recently. Instead of searching for an ``optimal'' adversarial example for a benign input to a targeted DNN, our algorithm finds a probability density distribution over a small region centered around the input, such that a sample drawn from this  distribution is likely an adversarial example, without the need of accessing the DNN's internal layers or weights. Our approach is \emph{universal} as it can successfully attack different neural networks by a single algorithm. It is also \emph{strong}; according to the testing against 2 vanilla DNNs and 13 defended ones, it outperforms state-of-the-art black-box or white-box attack methods for most test cases. Additionally, our results reveal that adversarial training remains one of the best defense techniques, and the adversarial examples are not as transferable across defended DNNs as them across vanilla DNNs.
\end{abstract}

\section{Introduction}

This paper is concerned with the robustness of deep neural networks (DNNs). We aim at providing a strong adversarial attack method that can universally defeat a variety of DNNs and associated defense techniques. Our experiments mainly focus on attacking the recently developed defense methods, following \cite{athalye2018obfuscated}. Unlike their work, however, we do not need to tailor our algorithm to various forms for tackling different defenses. Hence, it may generalize better to new defense methods in the future. Progress on powerful adversarial attack algorithms will significantly facilitate the research toward more robust DNNs that are deployed in uncertain or even adversarial environments.

\citet{szegedy2013intriguing} found that DNNs are  vulnerable to {adversarial examples} whose changes from the benign ones are imperceptible and yet can mislead DNNs to make wrong predictions. A rich line of work furthering their finding reveals more worrisome results. Notably, adversarial examples are \emph{transferable}, meaning that one can design adversarial examples for one DNN and then use them to fail others~\cite{papernot2016transferability,szegedy2013intriguing,tramer2017space}. Moreover, adversarial perturbation could be \emph{universal} in the sense that a single perturbation pattern may convert many images to adversarial ones~\cite{moosavi2017universal}. 

The adversarial examples raise a serious security issue as DNNs become increasingly popular~\cite{silver2016mastering,krizhevsky2012imagenet,hinton2012deep,Li_2018_ECCV,Gan_2017_ICCV}. Unfortunately, the cause of the adversarial examples remains unclear. \citet{goodfellow6572explaining} conjectured that DNNs behave linearly in the high dimensional input space, amplifying  small perturbations when their signs follow the DNNs' intrinsic linear weights. \citet{Fawzi_2018_CVPR} experimentally studied the topology and geometry of adversarial examples and \citet{DBLP:journals/corr/abs-1904-02057} provide the image-level interpretability
of adversarial examples. \citet{ma2018characterizing} characterized the subspace of adversarial examples. Nonetheless, defense methods~\cite{papernot2015distillation,tramer2017ensemble,rozsa2016towards,madry2017towards} motivated by them were  broken in a  short amount of time~\cite{he2017adversarial,athalye2018obfuscated,xu2017feature,sharma2017breaking}, indicating that better defense techniques are yet to be developed, and there may be unknown alternative factors that play a role in the DNNs' sensitivity.

Powerful adversarial attack methods are key to better understanding of the adversarial examples and for thorough testing of defense techniques. 

In this paper, we propose a black-box adversarial attack algorithm that can generate adversarial examples to defeat both vanilla DNNs and those recently defended by various techniques. Given an arbitrary input to a DNN, our algorithm finds a probability density over a small region centered around the input such that a sample drawn from this density distribution is likely an adversarial example, without the need of accessing the DNN's internal layers or weights --- thus, our method falls into the realm of black-box adversarial attack~\cite{papernot2017practical,brendel2017decision,chen2017zoo,ilyas2018black}.

Our approach is \emph{strong}; tested against two vanilla DNNs and 13 defended ones, it outperforms  state-of-the-art black-box or white-box attack methods for most cases, and it is on par with them for the remaining cases. It is also \emph{universal} as it attacks various DNNs by a single algorithm. We hope it can effectively benchmark new defense methods in the future --- code is available at {\small\url{https://github.com/Cold-Winter/Nattack}}.  Additionally, our study reveals that adversarial training remains one of the best defenses~\cite{madry2017towards}, and the adversarial examples are not as transferable across defended DNNs as them across  vanilla ones. The latter somehow weakens the practical significance of white-box methods which otherwise could fail a black-box DNN by attacking a substitute. 

Our optimization criterion is motivated by the natural evolution strategy (NES)~\cite{wierstra2008natural}. NES has been previously employed by~\citet{ilyas2018black} to estimate the gradients in the projected gradient search for adversarial examples. However, their algorithm leads to inferior performance  to what we proposed (cf.\ Table~\ref{table:resultdefense}). This is probably because, in their approach, the gradients have to be estimated relatively accurately for the projected gradient method to be effective. However, some of the neural networks $F(x)$ are not smooth, so that the NES estimation of the gradient $\nabla F(x)$ is not reliable enough.

In this paper, we opt for a different methodology using a constrained NES formulation as the objective function instead of using NES to estimate gradients as in~\citet{ilyas2018black}. The main idea is to smooth the loss function by a probability density distribution defined over the $\ell_p$-ball centered around a benign input to the neural network. All adversarial examples of this input belong to this ball\footnote{It is straightforward to extend our method to other  constraints bounding the offsets between inputs and adversarial examples.}. In this frame, assuming that we can find a distribution such that the loss is small, then a sample drawn from the distribution is likely adversarial. Notably, this formulation does not depend on estimating the gradient $\nabla F(x)$ any more, so it is not impeded by the non-smoothness of DNNs. 

We adopt parametric distributions in this work. The initialization to the distribution parameters plays a key role in the run time of our algorithm. In order to swiftly find a good initial distribution to start from, we train a regression neural network such that it takes as input the input to the target DNN to be attacked and its output parameterizes a probability density as the initialization to our main algorithm.

Our approach is advantageous over existing ones in multiple folds. First, we can designate the distribution in a \emph{low-dimensional} parameter space while the adversarial examples are often high-dimensional. Second, instead of questing an ``optimal'' adversarial example, we can virtually draw an \emph{infinite} number of adversarial examples from the distribution. Finally, the distribution may speed up the adversarial training for improving  DNNs' robustness because it is more \emph{efficient} to sample many  adversarial examples from a  distribution than to find them using gradient based optimization.

\section{Approach} \label{sApproach}

Consider a DNN classifier $C(x)=\arg\max_i F(x)_i$, where $x\in[0,1]^{\text{dim}(x)}$ is an input to the neural network $F(\cdot)$. We assume {softmax} is employed for the output layer of the network  and let $F(\cdot)_i$ denote the $i$-th dimension of the softmax output. When this DNN correctly classifies the input, {\it i.e.}, $C(x)=y$, where $y$ is the groundtruth label of the input $x$, our objective is to find an adversarial example $x_{adv}$ for $x$ such that they are imperceptibly close and yet the DNN classifier labels them distinctly; in other words, $C(x_{adv})\neq y$. {We exclude the inputs for which the DNN classifier predicts wrong labels in this work, following the convention of previous work~\cite{carlini2017towards}.}

We bound the $\ell_p$ distance  between an input $x$ and its adversarial counterparts: $x_{adv}\in S_p(x):=\{x':\|x-x'\|_p \leq \tau_p\}, p=2 \text{ or } \infty$. We omit from $S_p(x)$ the argument $(x)$ and the subscript $p$ when it does not cause ambiguity. Let $\proj_{S}(x')$ denote the projection of $x'\in\mathbb{R}^{\text{dim}(x)}$ onto $S$.

We first review the NES based black-box adversarial attack method~\cite{ilyas2018black}. We show that its performance is impeded by unstable estimation of the gradients of certain DNNs, followed by our approach which does not depend at all on the gradients of the DNNs.

\subsection{A Black-box Adversarial Attack by NES}

\citet{ilyas2018black} proposed to search for an optimal adversarial example in the following sense, 
\begin{align}
{x}_{adv} \;\leftarrow\; \arg\min_{x'\in S}\; f(x'), \label{eq:nes-prob}
\end{align}
given a benign input $x$ and its label $y$ correctly predicted by the neural network $F(\cdot)$, where $S$ is a small region containing $x$ defined above, and $f(x')$ is a loss function defined as $f(x'):=F(x')_y$. In  \cite{ilyas2018black}, this loss is minimized by the projected gradient method,
\begin{align}
    x_{t+1} \;\leftarrow\;\proj_S(x_t - \eta\, \sign(\nabla f(x_t))), \label{eq:nes-proj}
\end{align}
where $\sign(\cdot)$ is a sign function. The main challenge here is how to estimate the gradient $\nabla f(x_t)$ with derivative-free methods, as the network's internal architecture and weights are unknown in the black-box adversarial attack. One technique for doing so is by NES~\cite{wierstra2008natural}:
\begin{align}
\nabla f(x_t) &\approx \nabla_{x_t} \mathbb{E}_{\mathcal{N}(z|x_t,\sigma^2)}f(z) \label{eq:nes-gradient} \\ 
&= \mathbb{E}_{\mathcal{N}(z|x_t,\sigma^2)}f(z) \nabla_{x_t}\log\mathcal{N}(z|x_t,\sigma^2),
\end{align}
where $\mathcal{N}(z|x_t,\sigma^2)$ is an isometric normal distribution with mean $x_t$ and variance $\sigma^2$. Therefore, the stochastic gradient descent (SGD) version of eq.~(\ref{eq:nes-proj}) becomes: 
\begin{align}
x_{t+1}\leftarrow\proj_S\big(x_t-\eta\,\sign\big( \frac{1}{b}\sum_{i=1}^b f(z_i) \nabla \log \mathcal{N}(z_i|x_t,\sigma^2)\big)\big), \notag
\end{align}
where $b$ is the size of a mini-batch and $z_i$ is sampled from the normal distribution. The performance of this approach hinges on the quality of the estimated gradient. Our experiments show that its performance varies on attacking different DNNs probably because non-smooth DNNs lead to unstable NES estimation of the gradients (cf.\ eq.~(\ref{eq:nes-gradient})). 

\subsection{\ours}
We propose a different formulation albeit still motivated by NES. Given an input $x$ and a small region $S$ that contains $x$ (i.e., $S=S_p(x)$ defined earlier), the key idea is to consider a smoothed objective as our optimization criterion:
\begin{align}
    \min_\theta \; J(\theta):=\int f(x') \pi_S(x'|\theta) dx' \label{eq:our-prob}
\end{align}
where $\pi_S(x'|\theta)$ is a probability density with support defined on $S$. Compared with problem~(\ref{eq:nes-prob}), this frame assumes that we can find a distribution over $S$ such that the loss $f(x')$ is small in expectation. Hence, a sample drawn from this distribution is likely  adversarial. Furthermore, with  appropriate $\pi_S(\cdot|\theta)$, the objective $J(\theta)$ is a smooth function of $\theta$, and the optimization process of this formulation does not depend on any estimation of the gradient $\nabla f(x_t)$. Therefore, it is not impeded by the non-smoothness of neural networks. Finally, the distribution over $S$ can be parameterized in a much lower dimensional space ($\text{dim}(\theta)\ll\text{dim}(x)$), giving rise to more efficient algorithms than eq.~(\ref{eq:nes-proj}) which directly works in the high-dimensional input space.

\subsubsection{The distribution on $S$}
In order to define a distribution $\pi_S(x'|\theta)$ on $S$, we take the following transformation of variable approach:
\begin{align}
    x' = \proj_S(g(z)), \quad z\sim \mathcal{N}(z|\mu, \sigma^2)
\end{align}
where $\mathcal{N}(z|\mu, \sigma^2)$ is an isometric normal distribution whose mean $\mu$ and variance $\sigma^2$ are \emph{to be learned} and the function $g:\mathbb{R}^{\text{dim}(\mu)}\mapsto\mathbb{R}^{\text{dim}(x)}$ maps a normal instance to the space of the neural network input.  We leave it to future work to explore the other types of distributions. 

In this work, we implement the transformation of the normal variable by the following steps:
\begin{enumerate}
    \item draw $z\sim \mathcal{N}(\mu,\sigma^2)$, compute $g(z)$ as
\[
g(z)={1}/{2}(\tanh(g_0(z))+1),
\]
    \item clip $\delta'=\clip_p(g(z)-x)$,  $p=2 \text{ or } \infty$, and
    \item return $\proj_S(g(z))$ as  $x'=x+\delta'$
\end{enumerate}
Step 1 draws a ``seed'' $z$ and then maps it by $g_0(z)$ to the space of the same dimension as the input $x$. In our experiments,  we let $z$ lie in the space of the CIFAR10 images~\cite{cifar10} (i.e., $\mathbb{R}^{32\times 32\times 3}$), so the function $g_0(\cdot)$ is an identity mapping for the experiments on CIFAR10 and a bilinear interpolation for the ImageNet images~\cite{deng2009imagenet}. We further transform $g_0(z)$ to the same range as the input by $g(z)=\frac{1}{2}(\tanh{(g_0(z))}+1)\in[0,1]^{\text{dim}(x)}$ and then compute the offset $\delta=g(z)-x$ between the transformed vector and the input. Steps 2 and 3 detail how to project $g(z)$ onto the set $S$, where the clip functions are respectively
\begin{align}
\label{l2clip}
    \clip_2(\delta) &=\left\{
\begin{array}{cl}
\delta{\tau_2}/{\|\delta\|_2} & \text{if } \|\delta\|_2 > \tau_2 \\
\delta & \text{else}
\end{array}
\right. \\
    \clip_\infty(\delta) &= \min(\delta,\tau_\infty)
\end{align}
with the thresholds $\tau_2$ and $\tau_\infty$  given by users. 

Thus far, we have fully specified our problem formulation (eq.~(\ref{eq:our-prob})). Before discussing how to solve this problem, we recall that the set $S$ is the $\ell_p$-ball centered at $x$: $S=S_p(x)$.  Since problem (\ref{eq:our-prob}) is formulated for a particular input  to the targeted DNN, the input $x$ also determines the distribution $\pi_S(z|\theta)$ via the dependency of $S$ on $x$. In other words, we will learn personalized distributions for different inputs.

\subsubsection{Optimization}
Let $\proj_S(g(z))$ be steps 1--3 in the above variable transformation procedure. We can rewrite the objective function $J(\theta)$ in  problem~(\ref{eq:our-prob}) as 
\[
J(\theta) = \mathbb{E}_{\mathcal{N}(z|\mu,\sigma)} f(\proj(g(z))),
\]
where $\theta=(\mu,\sigma^2)$ are the unknowns. We use grid search to find a proper bandwidth $\sigma$ for the normal distribution  and NES to find its mean $\mu$:
\begin{align}
\mu_{t+1}\leftarrow\mu_t - \eta \nabla_\mu J(\theta)|_{\mu_t},
\end{align}
whose SGD version over a mini-batch of size $b$ is
\begin{align*}
\mu_{t+1}\leftarrow\mu_t - \frac{\eta}{b}\sum_{i=1}^b f(\proj_S(g(z_i)))\nabla_\mu\log \mathcal{N}(z_i|\mu_t,\sigma^2).
\end{align*}

In practice, we sample $\epsilon_i$ from a standard normal distribution and then use a linear transformation $z_i=\mu+\epsilon_i\sigma$ to make it follow the distribution $\mathcal{N}(z|\mu, \sigma^2)$. With this notion, we can simplify $\nabla_\mu\log \mathcal{N}(z_i|\mu_t,\sigma^2)\propto\sigma^{-1}\epsilon_i$. 

Algorithm~\ref{algorith} summarizes the full algorithm, called \ours, for optimizing our smoothed formulation in eq.~(\ref{eq:our-prob}). In line 6 of Algorithm~\ref{algorith}, the z-score operation is to subtract from each loss quantity $f_i$ the mean of the losses $f_1,\cdots,f_b$ and divide it by the standard deviation of all the loss quantities. We find it stablizes {\ours}; the algorithm converges well with a constant  learning rate $\eta$. Otherwise, one would have to schedule more sophisticated  learning rates as reported in~\cite{ilyas2018black}. Regarding the loss function in line 5, we employ the C\&W loss~\cite{carlini2017towards} in the experiments: $f(x'):=\max\big(0,\log F(x')_y - \max_{c\neq y}\log F(x')_c\big)$. 

In order to generate an adversarial example for an input $x$ to the neural network classifier $C(\cdot)$, we use the {\ours} algorithm to find a probability density distribution over $S_p(x)$ and then sample from this distribution until arriving at an adversarial instance $x'$ such that $C(x')\neq C(x)$.  

Note that our method differs from that of \citet{ilyas2018black} in that we allow an arbitrary data transformation $g(\cdot)$ which is more flexible than directly seeking the adversarial example in the input space, and we absorb the computation of $\proj_S(\cdot)$ into the function evaluation before the update of $\mu$ (line 7 of Algorithm~\ref{algorith}). On the contrary, the projection of \citet{ilyas2018black} is after the computation of the estimated gradient (which is similar to line 7 in Algorithm~\ref{algorith}) because it is an estimation of the projected gradient. The difference in the computational order of projection is conceptually important because, in our case, the projection is treated as part of the function evaluation, which is more stable than treating it as an estimation of the projected gradient. Practically, this also makes a major difference, which can be seen from our experimental comparisons of the two approaches.


\setlength{\textfloatsep}{10pt}
\begin{algorithm}[t]
\caption{Black-box adversarial  \ours}\label{euclid} \label{algorith}
\textbf{Input:} 
DNN  $F(\cdot)$, input $x$ and its label $y$, initial  mean ${\mu_0}$,  standard deviation ${\sigma}$, learning rate ${\eta}$, sample size $b$, and the maximum number of iterations $T$ \\
\textbf{Output:} $\mu_T$, mean of the normal distribution

\begin{algorithmic}[1]
  \FOR {$t = 0,1,...,T-1$} 
  \STATE Sample $\epsilon_1$,...,$\epsilon_b \sim  \mathcal{N}(0,I)$
  \STATE Compute $g_i = g(\mu_t + \epsilon_i\sigma)$ by Step 1 $\forall i \in \{1,\cdots,b\}$
  \STATE Obtain $\proj(g_i)$ by steps 2--3, $\forall i$
  \STATE  Compute losses $f_i := f(\proj(g_i)), \forall i$
  \STATE Z-score $\widehat{f}_i=(f_i-\text{mean}(f))/\text{std}(f), \forall i$
  \STATE Set $\mu_{t+1}\gets \mu_t -\frac{\eta}{b\sigma} \sum_{i=1}^b{\widehat{f}_i\epsilon_i}$
  \ENDFOR
\end{algorithmic}
\end{algorithm}

\subsection{Initializing {\ours} by Regression}
The initialization to the mean $\mu_0$ in Algorithm~\ref{algorith}  plays a key role in terms of run time. When a good initialization is given, we often successfully find adversarial examples in less than 100 iterations. Hence, we propose to boost the {\ours} algorithm by using a regression neural network. It takes a benign example $x$ as the input and outputs $\mu_0$ to initialize \ours. In order to train this regressor, we generate many (input, adversarial example) pairs $\{(x,x_{adv})\}$ by running {\ours} on the training set of benchmark datasets. The regression network's weights are then set by minimizing the $\ell_2$ loss between the network's output and $g_0^{-1}(\arctan(2 x_{adv}-1))-g_0^{-1}(\arctan(2 x-1))$; in other words, we regress for the offset between the adversarial example $x_{adv}$ and the input $x$ in the space $\mathbb{R}^{\text{dim}(\mu)}$ of the distribution parameters. The supplementary materials present more details about this regression network.

\section{Experiments}\label{exp}

We use the proposed {\ours} to attack \numdefense defense methods  for DNNs published in 2018 or 2019 and \numvanilla representative vanilla DNNs. For each defense method, we run experiments using the same protocol as reported in the original paper, including the datasets and $\ell_p$ distance (along with the threshold) to bound the differences between adversarial examples and inputs --- this experiment protocol favors the defense method. In particular, CIFAR10~\cite{cifar10} is employed in the attack on nine defense methods and ImageNet~\cite{deng2009imagenet} is used for the remaining four. We examine all the test images of CIFAR10 and randomly choose 1,000 images from the test set of ImageNet. 12 of the defenses concern the $\ell_\infty$ distance between the adversarial examples and the benign ones and one works with the $\ell_2$ distance.  We threshold the $l_\infty$ distance in the normalized $[0, 1]^{\text{dim}(x)}$ input space. The $l_2$ distance is normalized by the number of pixels.

In addition to the main comparison results, we also  investigate the defense methods' robustness versus the varying strengths of {\ours} (cf.\ Section~\ref{sec-curve}). Specifically, we plot the attack success rate versus the attack iteration. The curves provide a complementary metric to the overall attack success rate, uncovering the dynamic traits of the competition between a defense and an attack.  

Finally, we examine the adversarial examples' transferabilities between some of the \emph{defended neural networks} (cf.\ Section~\ref{sec-transfer}). Results show that, unlike the finding that many adversarial examples are transferable across different   \emph{vanilla neural networks}, a majority of the adversarial examples that fail one defended DNN cannot defeat the others. In some sense, this weakens the practical significance of white-box attack methods which are often thought applicable to unknown DNN classifiers by attacking a substitute neural network instead~\cite{papernot2017practical}.

\vspace{-10pt}
\subsection{Attacking \numdefense Most Recent Defense Techniques} \label{sec-comparison}

We consider \numdefense  defenses  recently developed: adversarial training (\textsc{Adv-train})~\cite{madry2017towards}, adversarial training of Bayesian DNNs (\textsc{Adv-BNN})~\cite{liu2018advbnn}, Thermometer encoding (\textsc{Therm})~\cite{buckman2018thermometer}, \textsc{Therm-Adv}~\cite{athalye2018obfuscated,madry2017towards}, \textsc{Adv-GAN}~\cite{wang2018a}, local intrinsic dimensionality (\textsc{LID})~\cite{ma2018characterizing},  stochastic activation pruning (\textsc{SAP})~\cite{dhillon2018stochastic}, random self-ensemble (\textsc{RSE})~\cite{liu2017towards}, cascade adversarial training (\textsc{Cas-adv})~\cite{na2018cascade},  randomization~\cite{xie2018mitigating}, input transformation (\textsc{Input-Trans})~\cite{guo2018countering}, pixel deflection~\cite{prakash2018deflecting}, and guided denoiser~\cite{liao2018defense}. We describe them in detail in the supplementary materials. Additionally, we also include  Wide Resnet-32 (\textsc{WResnet-32})~\cite{zagoruyko2016wide} and \textsc{Inception V3}~\cite{szegedy2016rethinking}, two vanilla neural networks for CIFAR10 and ImageNet, respectively.

{\bf Implementation Details.} In our experiments, the defended DNNs  of SAP, LID, \textsc{Randomization}, \textsc{Input-Trans}, \textsc{Therm}, and \textsc{Therm-dav} come from \cite{athalye2018obfuscated}, the defended models of  \textsc{Guided denoiser} and \textsc{Pixel deflection} are based on \cite{athalye2018robustness}, and the models defended by RSE, \textsc{Cas-adv}, \textsc{Adv-train}, and \textsc{Adv-GAN} are respectively from the original papers. For \textsc{Adv-BNN}, we attack an ensemble of ten BNN models. In all our experiments, we set $T=600$ as the maximum number of optimization iterations, $b=300$ for the sample size, variance of the isotropic Gaussian $\sigma^2 = 0.01$, and learning rate $\eta = 0.008$. {\ours} is able to defeat most of the defenses under this setting and about 90\% inputs for other cases. We then fine-tune the learning rate $\eta$ and sample size $b$ for the hard leftovers.

\begin{table*}[t]
\vspace{-7pt}
\centering
\caption{Adversarial attack on 13 recently published defense methods. (* the number reported in~\cite{athalye2018obfuscated}. For all the other numbers, we obtain them by running the code released by the authors or implemented ourselves with the help of the authors. For D-based and \textsc{Adv-Train},  we respectively report the results on 100 and 1000 images only because they incur expensive computation costs.)}
\label{table:resultdefense}
\resizebox{\columnwidth*2}{!}{
\begin{tabular}{llcl|ccccc} \toprule
\multirow{2}{*}{Defense Technique}                & \multirow{2}{*}{Dataset}    & Classification  & Threshold  & \multicolumn{4}{c}{Attack Success Rate \%}\\ 
&     & Accuracy \% & \& Distance &BPDA &ZOO & QL & D-based& \textbf{\ours}\\ 
\midrule
\textsc{Adv-train}
 & \multirow{2}{*}{CIFAR10} & \multirow{2}{*}{87.3} & \multirow{2}{*}{0.031 ($L_\infty$)} & \multirow{2}{*}{46.9} & \multirow{2}{*}{16.9}& \multirow{2}{*}{40.3}& \multirow{2}{*}{--}&\multirow{2}{*}{\textbf{47.9}}\\
 ~{\footnotesize \cite{madry2017towards}}       &        &      &&  & & &  \\ 
 
\textsc{adv-bnn} & \multirow{2}{*}{CIFAR10} & \multirow{2}{*}{79.7} &
\multirow{2}{*}{0.035 ($L_\infty$)} & \multirow{2}{*}{48.3} & \multirow{2}{*}{--}& \multirow{2}{*}{--} & \multirow{2}{*}{--} & \multirow{2}{*}{\textbf{75.3}}  \\
~{\footnotesize \cite{liu2018advbnn}}  &   &   &  & & & &  \\ 

\textsc{Therm-adv}
 & \multirow{2}{*}{CIFAR10} & \multirow{2}{*}{88.5} & \multirow{2}{*}{0.031 ($L_\infty$)} & \multirow{2}{*}{76.1}& \multirow{2}{*}{0.0}& \multirow{2}{*}{42.3}& \multirow{2}{*}{--}& \multirow{2}{*}{\textbf{91.2}}\\
 ~{\footnotesize \cite{athalye2018obfuscated}}       &        &      &  & & & & \\ 
 
 \textsc{Cas-adv} 
& \multirow{2}{*}{CIFAR10} & \multirow{2}{*}{75.6} &
\multirow{2}{*}{0.015 ($L_\infty$)} & \multirow{2}{*}{85.0*}  &  \multirow{2}{*}{96.1}  & \multirow{2}{*}{68.4}& \multirow{2}{*}{--} &\multirow{2}{*}{\textbf{97.7}}  \\
~{\footnotesize \cite{na2018cascade}}         &      &      &  & & &  & \\

 \textsc{ADV-GAN} 
& \multirow{2}{*}{CIFAR10} & \multirow{2}{*}{90.9} &
\multirow{2}{*}{0.031 ($L_\infty$)} & \multirow{2}{*}{48.9}  &  \multirow{2}{*}{76.4}  & \multirow{2}{*}{53.7}& \multirow{2}{*}{--} &\multirow{2}{*}{\textbf{98.3}}  \\
~{\footnotesize \cite{wang2018a}}         &      &      &  & & &  & \\ 

LID
& \multirow{2}{*}{CIFAR10} & \multirow{2}{*}{66.9} & \multirow{2}{*}{0.031 ($L_\infty$)} & \multirow{2}{*}{95.0}& \multirow{2}{*}{92.9} &\multirow{2}{*}{95.7}& \multirow{2}{*}{--}& \multirow{2}{*}{\textbf{100.0}}\\
~{\footnotesize \cite{ma2018characterizing}}       &        &      &  & & &  & \\ 
\textsc{Therm}
& \multirow{2}{*}{CIFAR10} & \multirow{2}{*}{92.8} & \multirow{2}{*}{0.031 ($L_\infty$)} & \multirow{2}{*}{\textbf{100.0}}& \multirow{2}{*}{0.0}& \multirow{2}{*}{96.5}& \multirow{2}{*}{--}& \multirow{2}{*}{\textbf{100.0}}\\
~{\footnotesize \cite{buckman2018thermometer}}     &          &      &  & & &  \\

SAP
& \multirow{2}{*}{CIFAR10} & \multirow{2}{*}{93.3} &
\multirow{2}{*}{0.031 ($L_\infty$)} & \multirow{2}{*}{\textbf{100.0}} & \multirow{2}{*}{5.9}& \multirow{2}{*}{96.2}& \multirow{2}{*}{--}&\multirow{2}{*}{\textbf{100.0}}\\
~{\footnotesize \cite{dhillon2018stochastic}}      &         &      &  & & & &  \\     

RSE & \multirow{2}{*}{CIFAR10} & \multirow{2}{*}{91.4} &
\multirow{2}{*}{0.031 ($L_\infty$)} & \multirow{2}{*}{--} & \multirow{2}{*}{--}& \multirow{2}{*}{--} & \multirow{2}{*}{--} & \multirow{2}{*}{\textbf{100.0}}  \\
~{\footnotesize \cite{liu2017towards}}  &   &   &  & & & &  \\

\textsc{\textit{VANILLA WRESNET-32}}  & \multirow{2}{*}{CIFAR10} & \multirow{2}{*}{95.0} &
\multirow{2}{*}{0.031 ($L_\infty$)} & \multirow{2}{*}{\textbf{100.0}} & \multirow{2}{*}{99.3}& \multirow{2}{*}{96.8} & \multirow{2}{*}{--} & \multirow{2}{*}{\textbf{100.0}}  \\
~{\footnotesize \cite{zagoruyko2016wide}}  &   &   &  & & & &  \\ 

\midrule

\textsc{Guided denoiser} & \multirow{2}{*}{ImageNet} & \multirow{2}{*}{79.1} &
\multirow{2}{*}{0.031 ($L_\infty$)} & \multirow{2}{*}{\textbf{100.0}} & \multirow{2}{*}{--} & \multirow{2}{*}{--}& \multirow{2}{*}{--}&  \multirow{2}{*}{95.5}  \\
~{\footnotesize \cite{liao2018defense}} &   &   &  & & & & \\ 

\textsc{Randomization} & \multirow{2}{*}{ImageNet} & \multirow{2}{*}{77.8} &
\multirow{2}{*}{0.031 ($L_\infty$)} & \multirow{2}{*}{\textbf{100.0}} & \multirow{2}{*}{6.7}& \multirow{2}{*}{45.9}& \multirow{2}{*}{--}& \multirow{2}{*}{96.5}  \\
~{\footnotesize \cite{xie2018mitigating}}         &      &      &  & & & & \\

\textsc{Input-Trans}  & \multirow{2}{*}{ImageNet} & \multirow{2}{*}{77.6} &
\multirow{2}{*}{0.05 ($L_2$)} & \multirow{2}{*}{\textbf{100.0}}  &  \multirow{2}{*}{38.3} & \multirow{2}{*}{66.5}& \multirow{2}{*}{66.0} & \multirow{2}{*}{\textbf{100.0}}  \\
~{\footnotesize \cite{guo2018countering}}               &  &    &  & & & & \\        

\textsc{Pixel deflection} & \multirow{2}{*}{ImageNet} & \multirow{2}{*}{69.1} &
\multirow{2}{*}{0.015 ($L_\infty$)} & \multirow{2}{*}{97.0} & \multirow{2}{*}{--}& \multirow{2}{*}{8.5}& \multirow{2}{*}{--}&  \multirow{2}{*}{\textbf{100.0}}  \\
~{\footnotesize \cite{prakash2018deflecting}}  &   &   &  & & & &  \\ 

\textit{{VANILLA INCEPTION V3}} & \multirow{2}{*}{ImageNet} & \multirow{2}{*}{78.0} &
\multirow{2}{*}{0.031 ($L_\infty$)} & \multirow{2}{*}{\textbf{100.0}} & \multirow{2}{*}{62.1}& \multirow{2}{*}{100.0}& \multirow{2}{*}{--}& \multirow{2}{*}{\textbf{100.0}} \\
~{\footnotesize \cite{szegedy2016rethinking}}         &      &      &  & & & & \\ 

\bottomrule
\end{tabular}
}
\vspace{-10pt}
\end{table*}

\vspace{-7pt}
\subsubsection{Attack success rates}
We report in Table~\ref{table:resultdefense} the main comparison results evaluated by the attack success rate, the higher the better. Our {\ours} achieves 100\% success on six out of the \numdefense defenses and more than 90\% on five of the rest. As a single black-box adversarial algorithm, {\ours} is  better than or on par with the set of powerful white-box attack methods of various forms~\cite{athalye2018obfuscated},  especially on the defended DNNs. It also significantly outperforms three state-of-the-art black-box attack methods: ZOO~\cite{chen2017zoo}, which adopts the zero-th order gradients to find  adversarial examples; QL~\cite{ilyas2018black}, a query-limited  attack based on an evolution strategy; and a decision-based (D-based) attack method~\cite{brendel2017decision} mainly generating $\ell_2$-bounded adversarial examples.  

Notably, {\textsc{Adv-train} is still among the best defense methods, so is its extension to the Bayesian DNNs (i.e., \textsc{Adv-BNN})}. However, along with  \textsc{Cas-Adv}  and \textsc{Therm-Adv} which are also equipped with the adversarial training, their strengths come at the price that they give  worse classification performances than the others on the clean inputs (cf.\ the third column of Table~\ref{table:resultdefense}). Moreover, \textsc{Adv-train} incurs extremely high computation cost. When the image resolutions are high, \citet{kurakin2016adversarial} found that it is difficult to run the adversarial training at the ImageNet scale. Since our \ours enables efficient generation of adversarial examples once we learn the distribution, we can potentially scale up the adversarial training with \ours and will explore it in the future work. 

We have tuned the main free parameters of the competing methods (e.g., batch size and bandwidth in QL). ZOO runs extremely slow with high-resolution images, so we instead use the hierarchical trick the authors described~\cite{chen2017zoo} for the experiments on ImageNet. In particular, we  run ZOO starting from the attack space of $32\times 32 \times 3$, lift the resolution to $64\times 64 \times 3$ after 2,000 iterations and then to $128 \times 128\times 3$ after 10,000 iterations, and finally up-sample the result to the same size as the DNN input with bilinear interpolation.  

\vspace{-7pt}
\subsubsection{Ablation study and run-time comparison}
\label{subsec-rumtime}
{\bf \ours vs.\ QL.} We have discussed the conceptual differences between \ours and QL~\cite{ilyas2018black} in Section~\ref{sApproach} (e.g., \ours formulates a smooth optimization criterion and offers a probability density on the $\ell_p$-ball of an input). Moreover, the comparison results in Table~\ref{table:resultdefense} verify the advantage of \ours over QL in terms of the overall attack strengths. Additionally, we here conduct an ablation study to investigate two major algorithmic differences between them: \ours absorbs the projection ($\proj_S$) into the objective function and allows an arbitrary change of variable transformation $g(\cdot)$. Our study concerns \textsc{Therm-Adv} and SAP, two defended DNNs on which QL respectively reaches 42.3\% and 96.2\% attack success rates. After we instead absorb the projection in QL into the objective, the results are improved to 54.7\% and 97.7\%, respectively. If we further apply $g(\cdot)$, the change of variable procedure (cf.\ Steps 1--3), the success rates become 83.3\% and 98.9\%, respectively. Finally, with  the z-score operation (line 6 of Algorithm~\ref{algorith}), the results are boosted to 90.9\%/100\%, approaching \ours's 91.2\%/100\%.  Therefore, we say that \ours boosts QL's performance, thanks to both the smoothed objective and the transformation $g(\cdot)$. 

{\bf \ours vs.\ the White-Box BPDA Attack.} While BPDA achieves high attack success rates  by different variants for handling the diverse defense techniques, \ours gives rise to better or comparable results by a single universal algorithm.  Additionally, we compare them in terms of the run time in the supplementary materials; the main observations are the following. On CIFAR10, BPDA and \ours can both find an adversarial example in about 30s. To defeat an ImageNet image, it takes {\ours} about 71s without the regression network and 48s when it is equipped with the regression net; in contrast, BPDA only needs 4s. It is surprising to see that BPDA is almost 7 times faster at attacking a  DNN for ImageNet than a DNN for CIFAR10. It is probably because the gradients of the former are not ``obfuscated'' as well as the latter due to the higher resolution of the ImageNet input.

\vspace{-10pt}
\subsection{Attack Success Rate vs.\ Attack Iteration }
\label{sec-curve}
\begin{figure*}[t]
  \centering
    \includegraphics[width = 1.0\textwidth]{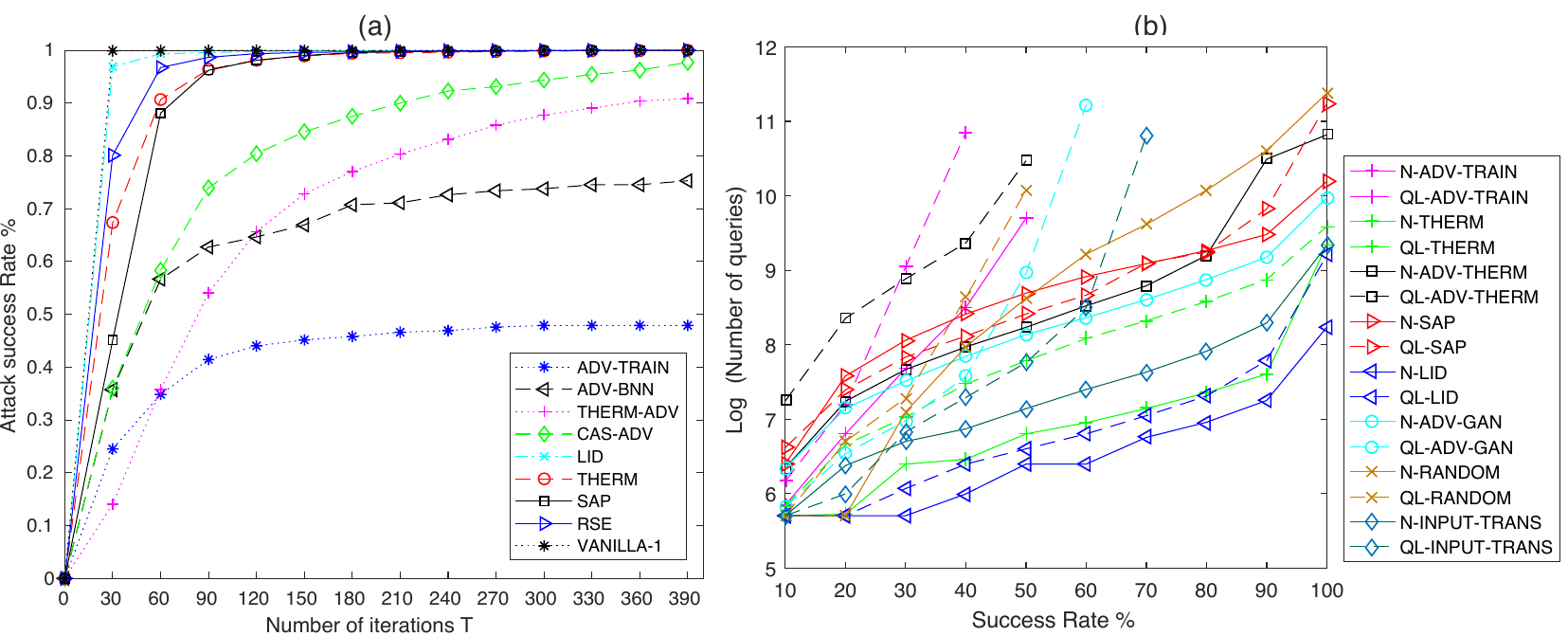}
    \vspace{-20pt}
  \caption{(a) Success rate versus run steps of \ours. (b) Comparison results with QL measured by the log of average number of queries per successful image. The solid lines denote \ours and the dashed lines illustrate QL.}
  \label{fig:runsteps}
   \vspace{-10pt}
\end{figure*}

The \ours algorithm has an appealing property as follows. In expectation, the loss (eq.~(\ref{eq:our-prob}))  decreases at every iteration and hence a sample  drawn from the distribution $\pi_S(x|\theta)$ is adversarial with higher chance. Though there could be oscillations, we find that the attack strengths do grow monotonically with respect to the evolution iterations in our experiments. Hence, we propose a new  curve shown in Figure~\ref{fig:runsteps}a featuring the attack success rate versus number of evolution iterations --- strength of attack. For the experiment here, the Gaussian mean  $\mu_0$ is initialized by $\mu_0\sim\mathcal{N}(g_0^{-1}(\arctan(2x-1)),\sigma^2)$ for any input $x$ to maintain about the same starting points for all the curves.

Figure~\ref{fig:runsteps}a plots eight defense methods on CIFAR10 along with a vanilla DNN. It is clear that \textsc{Adv-Train}, \textsc{Adv-BNN}, \textsc{Therm-Adv}, and \textsc{Cas-Adv}, which all employ the adversarial training strategy, are more difficult to attack than the others. What's more interesting is with the other five DNNs. Although {\ours} completely defeats them all  by the end, the curve of the vanilla DNN is the steepest while the \textsc{SAP} curve rises much slower. {If there are constraints on the computation time or the number of queries to the DNN classifiers, \textsc{SAP} is advantageous over the vanilla DNN, \textsc{RSE}, \textsc{Therm}, and \textsc{LID}}. 

Note that the ranking of the defenses in Table~\ref{table:resultdefense} (evaluation by the success rate) is different from the ordering on the left half of Figure~\ref{fig:runsteps}a, signifying the attack success rate and the curve mutually complement. The curve  reveals more characteristics of the defense methods especially when there are constraints on the computation time or number of queries to the DNN classifier. 

Figure~\ref{fig:runsteps}b shows \ours (solid lines) is more query efficient than the QL attack~\cite{ilyas2018black} (dashed lines) on 6 defenses under most attack success rates and the difference is even amplified for higher success rates. For SAP, \ours performs better when the desired attack success rate is bigger than $80\%$.



\vspace{-10pt}
\subsection{Transferability}
\label{sec-transfer}
We also study the transferability  of  adversarial examples across different \emph{defended} DNNs. This study differs from the earlier ones on \emph{vanilla} DNNs~\cite{szegedy2013intriguing,liu2016delving}. We investigate both the white-box attack BPDA and our black-box \ours.

Following the experiment setup in~\cite{kurakin2016adversarial}, we randomly select 1000 images for each targeted DNN such that they are classified correctly, and yet the adversarial images of them are classified incorrectly. We then use the adversarial examples of the 1000 images to attack the other DNNs.  In addition to the defended DNNs, we also include two vanilla DNNs for reference: \textsc{Vanilla-1} and \textsc{Vanilla-2}. \textsc{Vanilla-1} is a light-weight DNN classifier built by~\cite{carlini2017towards} with 80\% accuracy on CIFAR10. \textsc{Vanilla-2} is the Wide-ResNet-28~\cite{zagoruyko2016wide} which gives rise to 92.3\% classification accuracy on CIFAR10. For fair comparison, we change the threshold $\tau_\infty$ to 0.031 for  \textsc{Cas-adv}. We exclude RSE and \textsc{Cas-Adv} from BPDA's confusion table because it is not obviously clear how to attack RSE using BPDA and the released BPDA code lacks the piece for attacking \textsc{Cas-Adv}. 

\begin{figure*}[t]
  \centering
  \includegraphics[width = 1\textwidth]{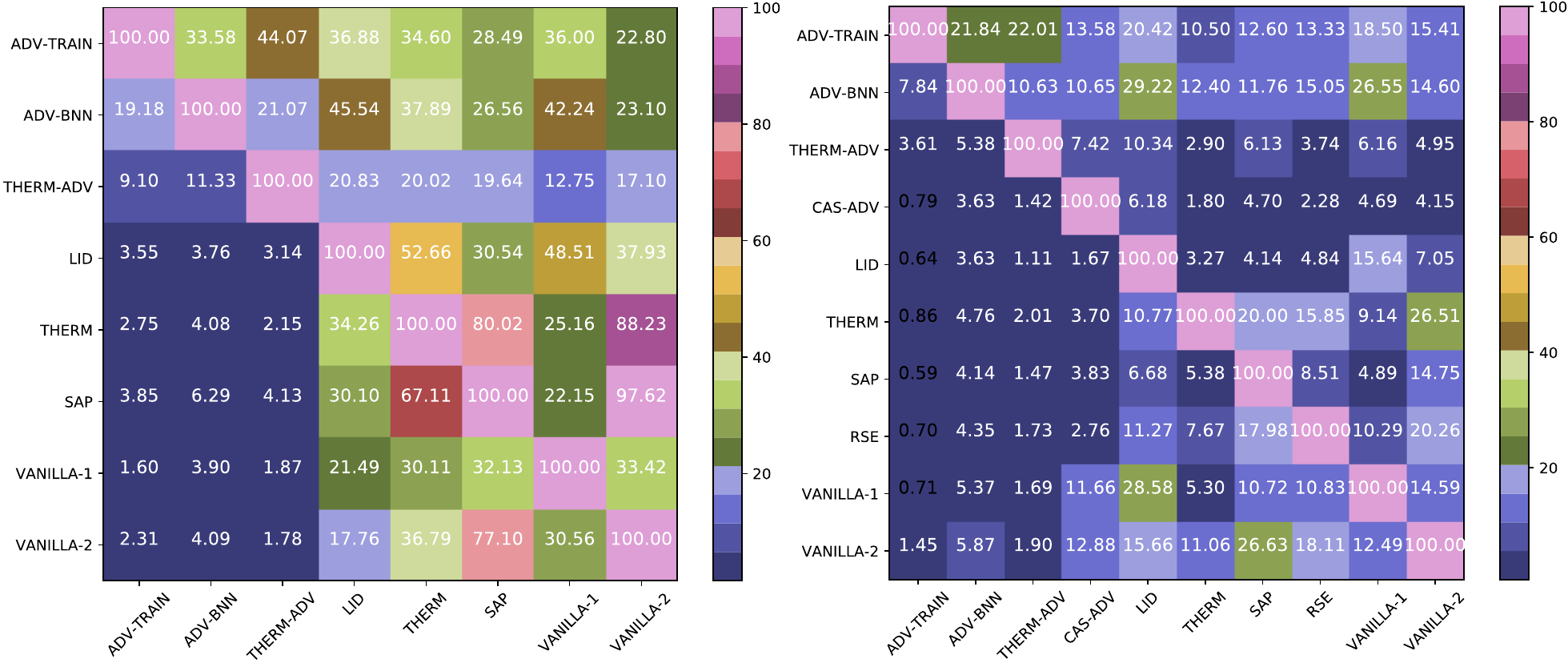}
   \vspace{-23pt}
  \caption{Transferabilities of BPDA~\cite{athalye2018obfuscated} (left) and {\ours} (right). Each entry shows the attack success rate of attacking the column-wise defense by the adversarial examples that are originally generated for the row-wise DNN.}
  \label{fig:confusion}
   \vspace{-12pt}
\end{figure*}

The confusion tables of BPDA and {\ours} are shown in Figure~\ref{fig:confusion}, respectively, where each entry indicates the success rate of using the adversarial examples originally targeting the row-wise  defense model to attack the column-wise defense. Both confusion tables are asymmetric; it is easier to transfer from defended models to the vanilla DNNs than vice versa. Besides, the overall transferrability is lower than that across the DNNs without any defenses~\cite{liu2016delving}. We highlight some additional observations below.

Firstly, the transferability of our black-box  {\ours} is not as good as the black-box BPDA attack. This is  probably because BPDA is able to explore the intrinsically common part of the DNN classifiers --- it has the privilege of accessing the true or estimated gradients that observe the  DNNs' architectures and weights. 

Secondly, both the network architecture and defense methods can influence the  transferability.  \textsc{Vanilla-2}  is the underlying classifier of SAP, \textsc{Therm-Adv}, and \textsc{Therm}. The adversarial examples originally attacking \textsc{Vanilla-2} do transfer better to SAP and \textsc{Therm} than to the others probably because they share the same DNN architecture, but the examples achieve very low success rate on \textsc{Therm-Adv}  due to the defense technique. 

Finally,  the transfer success rates are low no matter from \textsc{Therm-Adv} to the other defenses or vice versa, and \textsc{Adv-Train} and \textsc{Adv-BNN} lead to fairly good  results of transfer attacks on the other defenses and yet themselves are robust against the adversarial examples of the other defended DNNs. The unique result of \textsc{Therm-Adv} probably attributes to its use of double defense techniques, i.e., Thermometer encoding and  adversarial training.

\vspace{-10pt}
\section{Related Work}
\label{related}
There is a vast literature of adversarial attacks on and defenses for DNNs. We focus on the most related works in this section rather than a thorough survey. 

{\bf White-Box Attacks.} The adversary has full access to the target DNN in the white-box attack. \citet{szegedy2013intriguing} first find that DNNs are fragile to the adversarial examples by using box-constrained {L-BFGS}. \citet{goodfellow2014explaining} propose a fast gradient sign (\textit{FGS}) method, which is featured by efficiency and high performance for generating the $\ell_\infty$ bounded adversarial examples. \citet{papernot2016limitations} and \citet{moosavi2016deepfool} instead formulate the problems with the $l_0$ and $\ell_2$ metrics, respectively.  \cite{carlini2017towards} have proposed a powerful iterative optimization based attack. Similarly, a projected gradient descent has been shown strong in attacking DNNs~\cite{madry2017towards}. Most the white-box attacks rely on the gradients of the DNNs. When the gradients are ``obfuscated'' (e.g., by randomization), \cite{athalye2018obfuscated} derive various methods to approximate the gradients, while we use a single algorithm to attack a variety of defended DNNs. 

{\bf Black-Box Attacks.} As the name suggests, some parts of the DNNs are treated as black boxes in the black-box attack. Thanks to the adversarial examples' transferabilities~\cite{szegedy2013intriguing}, \citet{papernot2017practical} train a substitute DNN to imitate the target black-box DNN, produce adversarial examples of the substitute model, and then use them to attack the target DNN. \citet{chen2017zoo} instead use the zero-th order optimization to find adversarial examples. \citet{ilyas2018black} use the evolution strategy~\cite{salimans2017evolution} to approximate the gradients. \citet{brendel2017decision} introduce a decision-based attack by reading the hard labels predicted by a DNN, rather than the soft probabilistic output. Similarly, \citet{cheng2018queryefficient} also provide a formulation to explore the hard labels. Most of the existing black-box methods are tested against vanilla DNNs. In this work, we test them on defended ones along with our \ours.

\vspace{-10pt}
\section{Conclusion and Future Work}
\label{conclusion}
In this paper, we present a black-box adversarial attack method which learns a probability density on the $\ell_p$-ball of a clean input to the targeted neural network. One of the major advantages of our approach is that it allows an arbitrary transformation of variable $g(\cdot)$, converting the adversarial attack to a space of much lower dimensional than the input space. Experiments show that our algorithm defeats 13 defended  DNNs, better than or on par with  state-of-the-art white-box  attack methods. Additionally, our experiments on the transferability of the adversarial examples across the defended DNNs show different results reported in the literature: unlike the high transferability across vanilla DNNs, it is difficult to transfer the attacks on the defended DNNs.


Some existing works try to characterize the adversarial examples by their geometric properties. In contrast to this macro view, we model the adversarial population of each single input from a micro view by a probabilistic density. There are still a lot to explore along this avenue.  \emph{What is a good family of distributions to model the adversarial examples? How to conduct adversarial training by efficiently sampling from the distribution?} These questions are worth further investigation in the future work.  

\paragraph{Acknowledgement:} This work was supported in part by NSF-1836881, NSF-1741431, and ONR-N00014-18-1-2121.

\bibliographystyle{icml2019}



\appendix
\newpage


\def \ours {$\mathcal{N}$\textsc{Attack}}


\icmltitlerunning{\ours: Learning the Distributions of Adversarial Examples for an Improved  Black-Box  Attack on Deep Neural Networks}


\twocolumn[
\icmltitle{Supplementary Materials for \\ \ours: Learning the Distributions of Adversarial Examples for an Improved  Black-Box  Attack on Deep Neural Networks}

\vskip 0.3in
]




In this supplementary document, we provide  the following details to support the main text:
\begin{description}
\item[Section~\ref{app-defenses}:] descriptions of the 13 defense methods studied in the experiments,
\item[Section~\ref{app-regression}:] architecture of the regression neural network for initializing our {\ours} algorithm, and
\item[Section~\ref{appendix-runtime}:] run-time analysis about {\ours} and BPDA~\cite{athalye2018obfuscated}.
\end{description}

\section{More Details of the 13 Defense Methods} \label{app-defenses}

\begin{itemize}[leftmargin=*]

\item {\bf Thermometer encoding (\textsc{Therm}).} To break the hypothesized linearity behavior of DNNs~\citep{goodfellow2014explaining}, \citet{buckman2018thermometer} proposed to transform the input by non-differentiable and non-linear thermometer encoding, followed by a slight change to the input layer of conventional DNNs.

\item {\bf \textsc{Adv-Train \& Therm-Adv}.}
\citet{madry2017towards} proposed a defense using adversarial training (\textsc{Adv-Train}). Specially, the training procedure alternates between seeking an ``optimal'' adversarial example for each input by projected gradient descent (PGD) and minimizing the classification loss under the PGD attack. Furthermore, \citet{athalye2018obfuscated} find that the adversarial robust training~\cite{madry2017towards} can significantly improve the defense strength of \textsc{Therm} (\textsc{Therm-Adv}). Compared with \textsc{Adv-Train}, the adversarial examples are produced by the logit-space projected gradient ascent in the training.

\begin{table*}[htbp]
\centering
\caption{Average run time to find an adversarial example (\textbf{\ours-R} stands for \ours~initialized with the regression net).}
\label{table:Avgtime}
\begin{tabular}{lllll}
\toprule
 \multirow{2}{*}{Defense} & \multirow{2}{*}{Dataset} & BPDA & \multirow{2}{*}{\ours}  &\multirow{2}{*}{\textbf{\ours-R}}\\

& &~\cite{athalye2018obfuscated} & \\
\midrule
 SAP & \multirow{2}{*}{CIFAR-10 ($L_\infty$)} & \multirow{2}{*}{33.3s} & \multirow{2}{*}{29.4s} & \multirow{2}{*}{--}\\
 ~{\footnotesize \cite{dhillon2018stochastic}} & & & &
 \\
 
\textsc{Randomization} &\multirow{2}{*}{ImageNet ($L_\infty$)} & \multirow{2}{*}{3.51s}  & \multirow{2}{*}{70.77s} & \multirow{2}{*}{48.22s}\\
~{\footnotesize \cite{xie2018mitigating}} & & & & \\
\bottomrule
\end{tabular}
\end{table*}

\item {\bf Cascade adversarial training (\textsc{Cas-adv}).} \citet{na2018cascade} reduced the computation cost of the adversarial training~\citep{goodfellow6572explaining,kurakin2016adversarial}
in a cascade manner. A model is trained from the clean data and one-step adversarial examples first. The second model is trained from the original data, one-step adversarial examples, as well as iterative adversarial examples generated against the first model. Additionally, a regularization is introduced to the unified embeddings of the clean and adversarial examples. 

\item {\bf Adversarially trained Bayesian neural network (\textsc{ADV-BNN}).}
\citet{liu2018advbnn} proposed to model the randomness added to DNNs in a Bayesian framework in order to defend against adversarial attack. Besides, they incorporated the adversarial training, which has been shown effective in the previous works, into the framework.  

\item {\bf Adversarial training with adversarial examples generated from GAN (\textsc{ADV-GAN}).} \citet{wang2018a} proposed to model the adversarial perturbation with a generative network, and they learned it jointly with the defensive DNN as a discriminator.

\item {\bf Stochastic activation pruning (\textsc{SAP}).} \citet{dhillon2018stochastic} randomly dropped
some neurons of each layer with the probabilities in proportion
to their absolute values. 

\item {\bf \textsc{Randomization}.}~\cite{xie2018mitigating} added a  randomization
layer between inputs and a DNN classifier. This layer consists of resizing an image to a random resolution, zero-padding, and randomly selecting one from many resulting images as the actual input to the classifier. 

\item {\bf {Input transformation} (\textsc{Input-Trans}).} By a similar idea as above, \citet{guo2018countering} explored several combinations of input transformations coupled with adversarial training, such as image cropping and rescaling, bit-depth reduction, JPEG compression. 

\item {\bf \textsc{Pixel deflection}.} \citet{prakash2018deflecting} randomly sample a pixel from an image and then replace it with another pixel randomly sampled from the former's neighborhood. Discrete wavelet transform is also employed to filter  out adversarial perturbations to the input.

\item {\bf \textsc{Guided denoiser}.} \citet{liao2018defense} use a denoising network architecture to estimate the additive adversarial perturbation to an input. 

\item {\bf Random self-ensemble (\textsc{RSE}).} \citet{liu2017towards} combine the ideas of randomness and ensemble using the same underlying neural network. Given an input, it generates an ensemble of predictions by adding distinct noises to the network multiple times. 

\end{itemize}

\section{Architecture of the Regression Network} \label{app-regression}
We construct our regression neural network by using the fully convolutional network (FCN) architecture~\cite{shelhamer2016fully}. In particular, we adapt the FCN model pretrained on PASCAL VOC segmentation challenge~\citep{Everingham10} to our work by changing its last two layers, such that the network outputs an adversarial perturbation of the size $32\times 32\times 3$. We train this network by a mean square loss. 



 




 

\section{Run Time Comparison} \label{appendix-runtime}
Compared with the white-box attack approach BPDA~\citep{athalye2018obfuscated}, \ours~may take longer time since BPDA can find the local optimal solution quickly being guided by the approximate gradients. However, {\ours} can be executed in parallel in each episode. We leave implement the parallel version of our algorithm to the future work and compare its sing-thread version with BPDA below. 

We attack 100 samples on one machine with fou TITAN-XP graphic cards and calculate the average run time for reaching an adversarial example. As shown in Table~\ref{table:Avgtime}, \ours~can succeed even faster than the white-box BPDA on CIFAR-10, yet runs slower on ImageNet.
The main reason is that when the image size is as small as CIFAR10 (3*32*32), the search space is moderate. However, the run time could be lengthy for high resolution images like ImageNet  (3*299*299) especially for some hard cases (we can find the adversarial examples for nearly 90\% test images but it could take about 60 minutes for a hard case).

We use a regression net to approximate a good initialization of $\mu_0$ and we name \ours~initialized with the regression net as \ours-R. We run \ours~and \ours-R on ImageNet with the mini-batch size $b=40$ . The success rate for \ours~with random initialization is 82\% and for \ours-R is 91.9\%, verifying the efficacy of the regression net. The run time shown in Table~\ref{table:Avgtime} is calculated on the images with successful attacks. The results demonstrate that \ours-R can reduce by 22.5s attack time per image compared with the random initialization.

\end{document}